# A Survey of Fish Tracking Techniques Based on Computer Vision

Weiran Li[a,b,c,d,e], Zhenbo Li[a,b,c,d,e,*], Fei Li[a,b,c,d,e], Meng Yuan[a], Chaojun Cen[a,b,c,d,e], Yanyu Qi[a,b,c,d,e], Qiannan Guo[a,b,c,d,e], You Li[f]

[a]*College of Information and Electrical Engineering, China Agricultural University, Beijing 100083, China*

[b]*National Innovation Center for Digital Fishery, Ministry of Agriculture and Rural Affairs, Beijing 100083, China*

[c]*Key Laboratory of Agricultural Information Acquisition Technology, Ministry of Agriculture and Rural Affairs, Beijing 100083, China*

[d]*Beijing Engineering and Technology Research Center for Internet of Things in Agriculture, Beijing 100083, China*

[e]*Key Laboratory of Smart Farming for Aquatic Animal and Livestock, Ministry of Agriculture and Rural Affairs, Beijing 100083, China*

[f]*College of Engineering, China Agricultural University, Beijing 100083, China*

**Abstract:** Fish tracking is a key technology for obtaining movement trajectories and identifying abnormal behavior. However, it faces considerable challenges, including occlusion, multi-scale tracking, and fish deformation. Notably, extant reviews have focused more on behavioral analysis rather than providing a comprehensive overview of computer vision-based fish tracking approaches. This paper presents a comprehensive review of the advancements of fish tracking technologies over the past seven years (2017-2023). It explores diverse fish tracking techniques with an emphasis on fundamental localization and tracking methods. Auxiliary plugins commonly integrated into fish tracking systems, such as underwater image enhancement and re-identification, are also examined. Additionally, this paper summarizes open-source datasets, evaluation metrics, challenges, and applications in fish tracking research. Finally, a comprehensive discussion offers insights and future directions for vision-based fish tracking techniques. We hope that our work could provide a partial reference in the development of fish tracking algorithms.

## 1 Introduction

According to the Food and Agriculture Organization (FAO) report, there is a projected global increase in aquatic output, with expectations to reach 204 million tons by the year 2030 (Stankus, 2021). Fish farming is a pivotal component of aquaculture, constituting approximately 80 percent of total aquaculture production. The application of fish tracking technology is of paramount importance for the advancement and enhancement of aquaculture practices, ultimately resulting in increased efficiency and smartification of the industry.

Fish tracking technology is undergoing a transformation, where the tracking paradigm is shifting from being dominated by a manual workflow to an end-to-end deep learning paradigm (Yang et al., 2021). Fish identification and tracking methods play a critical role in marine monitoring, aquaculture and biological research by enabling functions such as disease identification and hypoxic stress detection, which have significant implications for the fisheries sector. In contrast to techniques relying on acoustics or sensors, fish tracking methods based on computer vision offer several advantages such as real-time monitoring, non-invasiveness, minimal equipment requirements, and the preservation of natural fish behavior (D. Li et al., 2020). In addition, images contain rich semantic information that can be leveraged by deep networks to enable other downstream tasks such as segmentation and behavioral analysis.

* Corresponding author at: P.O. Box121, China Agricultural University, 17 Tsinghua East Road, Beijing 100083, PR China. E-mail address: lizb@cau.edu.cn (Z.Li).

Fish object tracking involves predicting fish trajectories within video data. Fundamental tracking modules such as the Kalman filter, particle filter, and other generative models are extensively utilized in fish tracking (Nian et al., 2017; Zhao et al., 2019). A common approach in fish tracking methods combines background subtraction with the Kalman filter (Eldrogi, 2019; França Albuquerque et al., 2019; Zhao et al., 2019). Additionally, the field of multiple fish tracking is currently exploring two prominent paradigms: Separate Detection and Embedding (SDE) and Joint Detection and Embedding (JDE). Both paradigms are based on deep learning techniques (Li et al., 2022; Zhao et al., 2019).

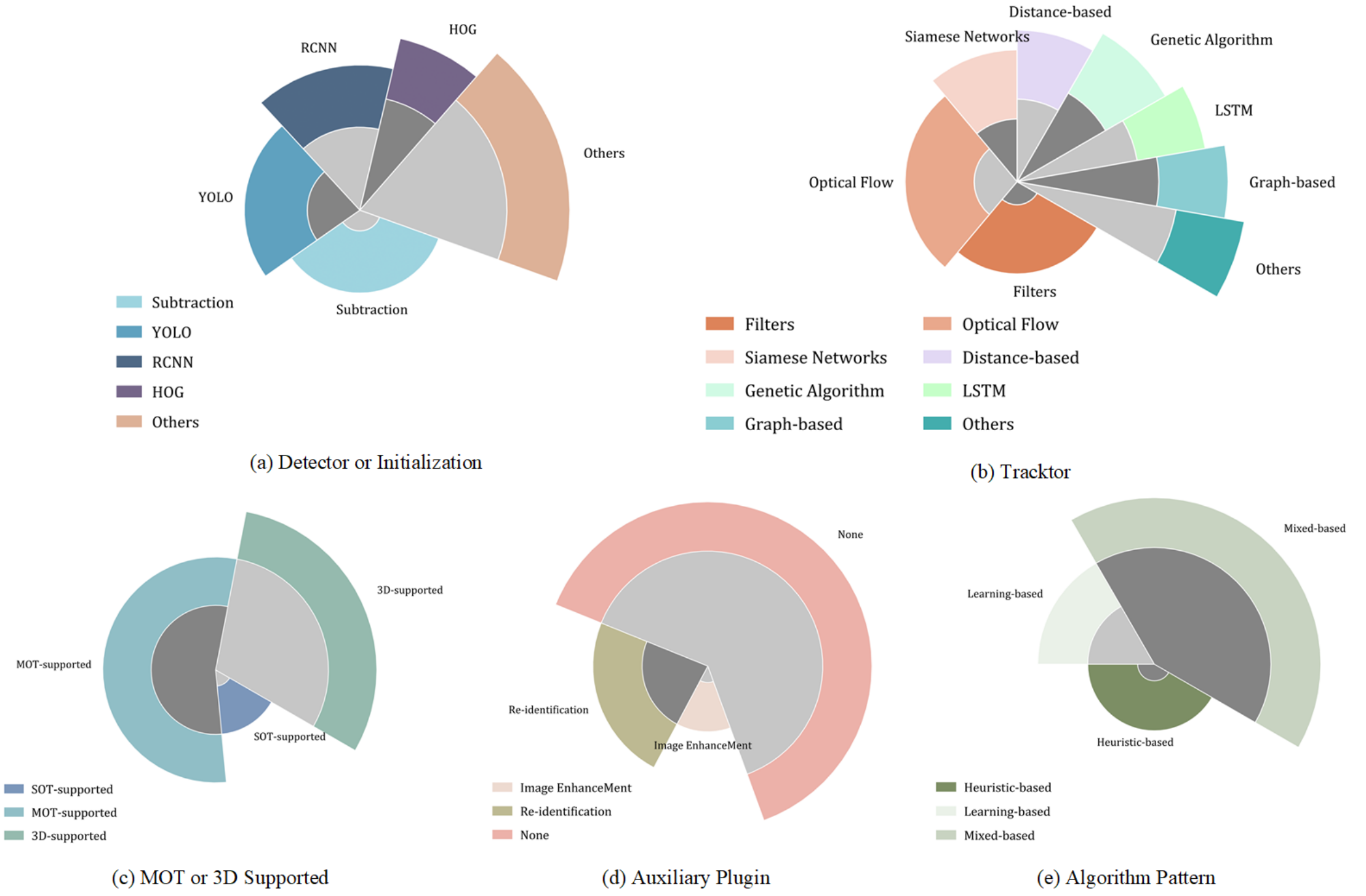

(a) Detector or Initialization (b) Tracktor

(c) MOT or 3D Supported (d) Auxiliary Plugin (e) Algorithm Pattern

**Fig. 1** Fish tracking methods in recent 7 years (2017~2023). Based on our prior knowledge, these methods are classified using the following five criteria: 1) Detector or Initialization (Fig. 1a). For Multiple Object Tracking (MOT) and Single Object Tracking (SOT), the detection and initialization methods are adopted as the division basis, respectively. 2) Tracktor (Fig. 1b). It includes the association methods and inter-frame calculation methods both in MOT and SOT. 3) MOT or 3D Supported (Fig. 1c). It is divided according to whether the algorithm supports multi-target tracking and 3D tracking. The priority classification of 3D tracking is higher than that of MOT, because most 3D methods usually support multi-target. 4) Auxiliary Plugin (Fig. 1d). This item is divided according to whether the tracking method contains other auxiliary plugins. 5) Algorithm Pattern (Fig. 1e). The division is based on whether the method relies on data-driven.

In our comprehensive literature review, we have categorized fish tracking methods from the past seven years (2017-2023) into different categories, as shown in Fig. 1. The findings suggest that traditional methodologies often exhibit suboptimal performance when faced with the challenges of underwater environments (Mohamed et al., 2020). In contrast, tracking technologies based on deep learning (Liu et al., 2021; H. Wang et al., 2022) excel at adapting to the characteristics of underwater data distribution and make effective use of extensive datasets to train versatile models. Deep learning-based tracking algorithms have found extensive applications in domains such as pedestrian and vehicle tracking, demonstrating their versatility. These algorithms employ a discriminative tracking process, effectively leveraging background information to distinguish and identify targets. It is worth noting that the adoption of deep learning-based tracking methods in contexts related to fish is still relatively limited.

The structural organization of this survey is illustrated as Fig. 2, with the following sections presented as follows: Section 2 provides an introduction to the characteristics of fish tracking algorithms, including heuristic-based, learning-based, and mixed-based approaches, along with the supported tracking targets or dimensions. Section 3 provides an in-depth look at the basic methods used in fish tracking algorithms, with particular emphasis on localization and tracking techniques. In Section 4, we explore common auxiliary plugins, including underwater image enhancement modules and Re-ID modules. Section 5 offers insights into open-source datasets and commonly employed tracking metrics in fish tracking. Section 6 gives an overview of challenges and applications. Finally, a discussion is presented in Section 7, and conclusion of the survey are summarized in Section 8.

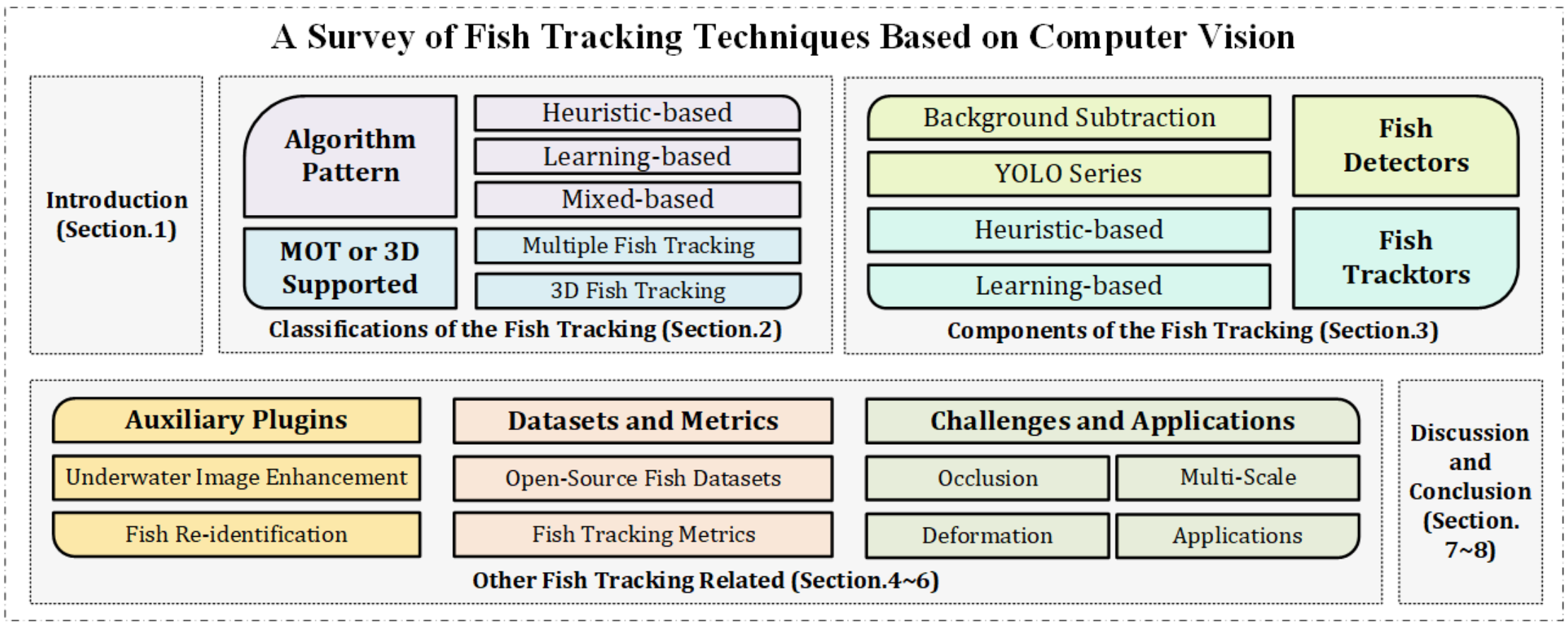

**Fig. 2** Contents and structural organization of the survey.

# 2 Categories of the Fish Tracking

The various fish tracking algorithms are applied for different purposes and therefore vary in the number of targets supported, accuracy, and real-time performance (França Albuquerque et al., 2019; Mohamed et al., 2020). For example, in biological studies where the goal is to track individual fish and obtain more accurate trajectories, offline tracking of single targets is often relied upon. In contrast, aquaculture monitoring prioritizes faster online tracking methods for multiple fish targets, with the aim of assessing the normal physiological state of the fish population. In this section, we categorize fish tracking algorithms based on their architectural characteristics, number of supported targets, and dimensions supported.

## 2.1 Algorithm Pattern

With the widespread adoption of deep learning algorithms, fish tracking algorithms can be categorized into three distinct groups based on whether the model utilizes a data-driven approach (Fig. 1e): Heuristic-based methods, Learning-based methods, and Mixed-based methods. Fig. 3 presents examples of typical frameworks representing various fish tracking patterns.

Heuristic-based methods refer to models designed without any data-driven training process, making them generalizable to various scenarios directly. Traditional fish tracking algorithms, especially in the detection, initialization, and tracking stages, often rely on heuristic-based methods (Eldrogi, 2019; Nian et al., 2017). Fig. 3a illustrates a typical processing framework in this category. For instance, Zhao et al. (2019) devised a tracking algorithm for red snapper, aimed at mitigating occlusion within fish schools due to water quality issues. Their approach utilizes the Otsu adaptive segmentation algorithm to extract fish targets through background subtraction and employs the Kalman filter for motion estimation. Similarly, Lumauag and Nava (2018) presented a fish tracking method based on image processing, utilizing blob analysis and Euclidean filter for tracking and counting. However, it is worth noting that achieving generic fish tracking in underwater scenarios using a single unified heuristic model can be challenging due to domain-specific differences.

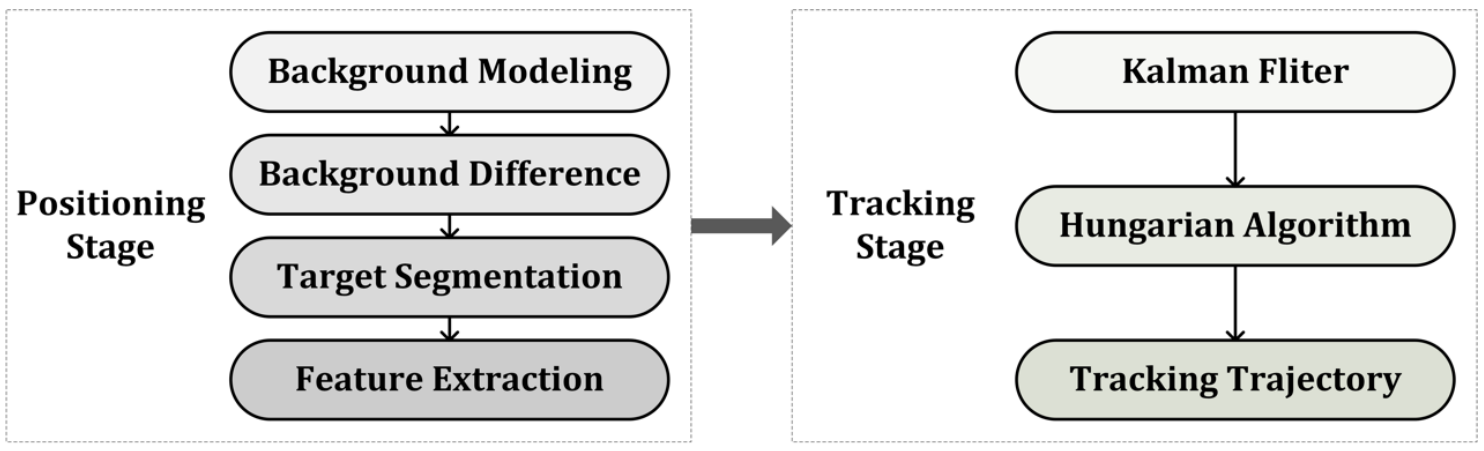

(a) Heuristic-based

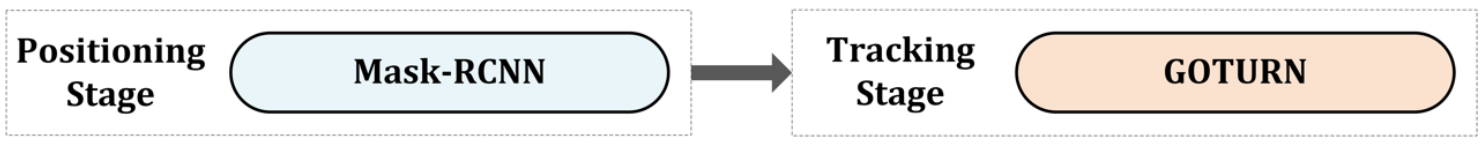

(b) Learning-based

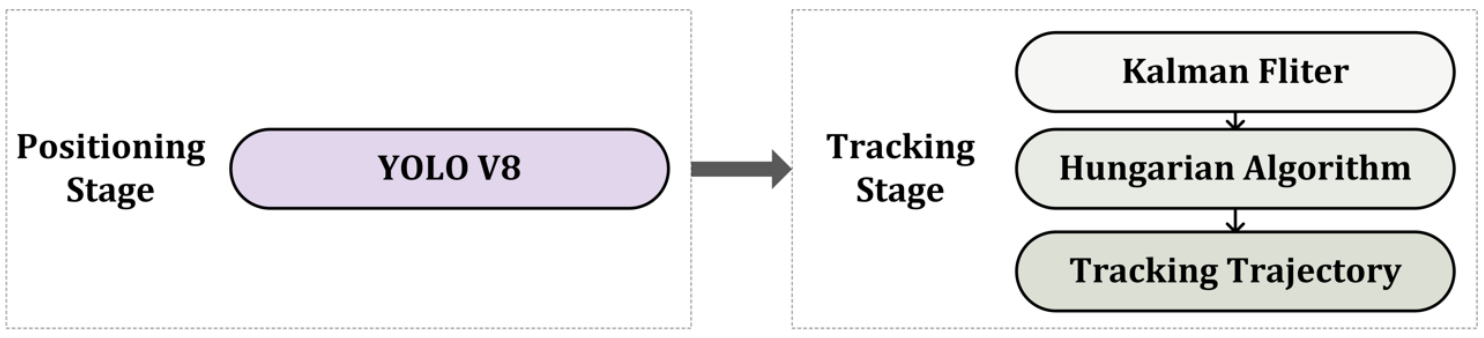

(c) Mixed-based

Fig. 3 Framework examples of various fish tracking patterns.

Learning-based methods incorporate either a unified learnable network or two separate networks for detection and association. Fig. 3b illustrates a typical framework that includes a detection method like the RCNN series (He et al., 2017) and GOTURN (Held et al., 2016). For instance, Arvind et al. (2019) introduced a real-time fish detection and tracking approach based on Mask RCNN (He et al., 2017) and GOTURN. They utilized a UAV-mounted vision sensor to capture fish images in murky water, employing Mask RCNN for fish segmentation and GOTURN for tracking. This method achieved a frame rate of 16 frames per second (FPS) in inland fish environments. While deep learning-based holistic fish tracking frameworks can provide compatibility across varied scenes, the model training process tends to be more complex and may lead to relatively poorer real-time performance compared to traditional heuristic tracking models.

Mixed-based methods involve a framework that integrates both heuristic-based and learnable-based components. Typically, deep learning-based detectors like the YOLO series (C.-Y. Wang et al., 2022) are used in the detection phase, while traditional heuristics like SORT (Bewley et al., 2016) are applied in the tracking phase. This framework is depicted in Fig. 3c. For instance, Wageeh et al. (2021) introduced a fish tracking approach that utilizes MSR-YOLO and Euclidean distance, effectively integrating learning-based and heuristic-based components. The improved YOLO algorithm is employed for fish detection and coordinate positioning, while simple Euclidean distance calculations between frames are used for instance ID assignment. This method achieves accurate fish tracking in low-light underwater scenarios. Mixed-based models usually demonstrate superior detection capabilities along with higher real-time performance, making them promising candidates for aquaculture applications.

## 2.2 MOT or 3D Supported

2D-based single-target tracking often has limited applicability. On one hand, there are few aquaculture scenarios where individual fish are isolated. On the other hand, obtaining their spatial coordinates for trajectory tracking can be challenging, which imposes constraints for biological studies. Therefore, multiple fish tracking and 3D fish tracking are better positioned to satisfy the practical requirements of researchers and breeders.

Multiple Object Tracking (MOT) methods refer to tracking models capable of simultaneously predicting trajectories for multiple targets, facilitating the mutual identification of individual instances between consecutive frames. Most mainstream methods employ a Tracking-By-Detection (TBD) paradigm to track fish targets (Liu et al., 2018; Wageeh et al., 2021), as illustrated in Fig. 4. In this paradigm, the current state of the target serves as input to predict its position in the next frame, and association algorithms are then employed to make corrections. For instance, Shreesha et al. (2023) introduced a multiple fish tracking method based on YOLOv3 and a cost matrix. YOLOv3 serves as the detector to extract the positions of various fish instances, while the cost matrix is utilized for assigning trajectories to corresponding fish instances. Additionally, JDE-based fish tracking methods such as CMFTNet (Li et al., 2022) can achieve high-performance online multiple fish tracking with relatively end-to-end model training. Fish tracking algorithms built upon MOT principles have exhibited remarkable capabilities in aquaculture applications and have become the dominant design paradigm.

3D-based tracking methods offer superior performance in biological studies due to their precise spatial information and effective handling of scene occlusion. Typically, these methods employ multiple cameras to capture simultaneous shots, and the 3D coordinate information of the target is obtained through alignment calculations. For instance, Palconit et al. (2021) introduced a fish tracking method based on stereo vision, which can effectively address the rapid movements of fish. Two cameras are used to capture videos, and a triangulation method is utilized to generate the z-coordinate of fish images. The 3D tracking method can more accurately determine fish behavior and detect abnormalities compared to 2D tracking methods. However, the application of 3D tracking is limited due to hardware constraints and the high computational complexity of image alignment and scene modeling for real-world environments.

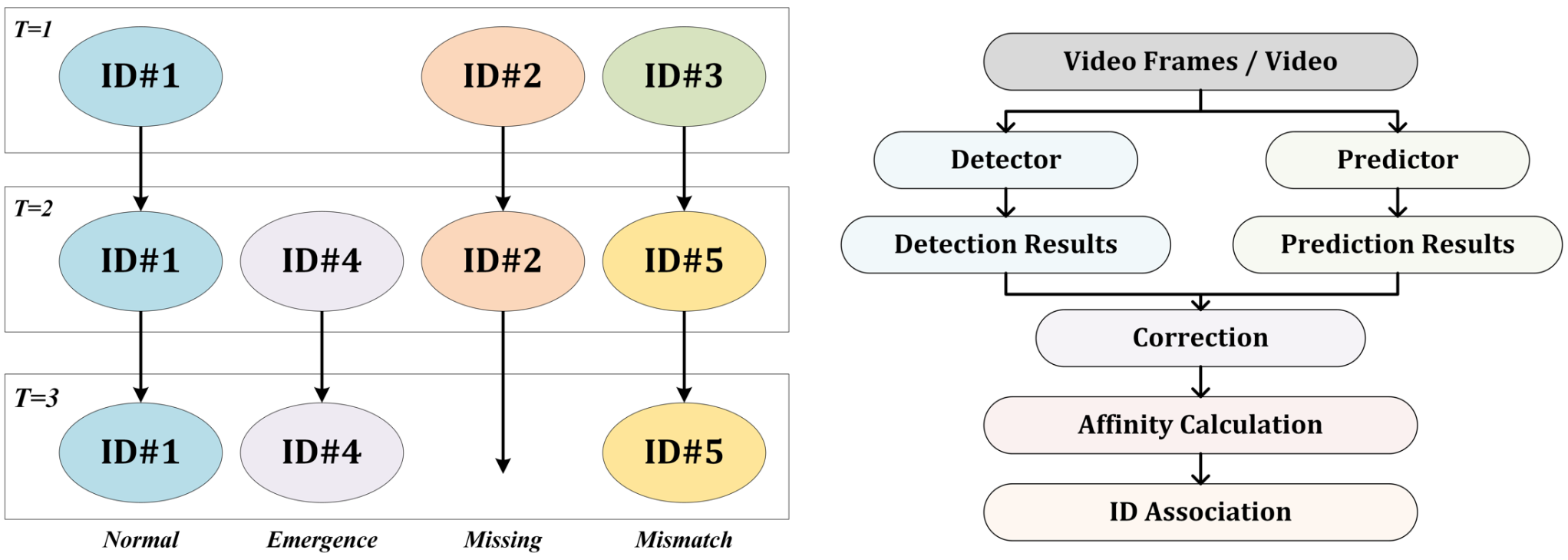

(a) Multiple Fish Tracking Inter-frame Processing (b) Multiple Fish Tracking General Processing

Fig. 4 Processing details of multiple fish tracking. The Tracking-By-Detection (TBD) paradigm is taken here as an example.

## 2.3 Others

In the context of tracking methods applied to targets other than fish, various classification benchmarks exist. However, when focusing on fish targets, challenges often arise due to dataset-specific constraints such as rapid fish movements and frequent occlusion occurrences. As a result, some algorithms may not have been extensively tested for tracking fish targets. To facilitate further research, this section introduces alternative classification criteria beyond the divisions mentioned above. The criteria for these different divisions and their representation methods are summarized in Table 1.

**Table 1.** Various classification bases of tracking algorithms.

| Classification | Method | Introduction | Example |
|---|---|---|---|
| **Task Calculation** | Online | Real-time processing of tasks, using only current and past frames to track the position of objects on future frames | (Guo et al., 2017) |
| | Offline | Offline processing tasks, using past, present, and future frames for object position tracking, with high accuracy | (Guo et al., 2021) |
| **Tracking Task** | SOT | Track the location of a given target | (Bertinetto et al., 2016) |
| | MOT | Track the location of multiple targets | (Zhou et al., 2020) |
| | Re-ID | Considered as a sub-problem of image retrieval, judging the similarity with a given picture | (Zheng et al., 2021) |
| | MTMCT | Multi-target multi-camera tracking, considered as is an extension of Re-ID | (Ristani and Tomasi, 2018) |
| **Method Category** | Generative | Establish a target model or extract target features, search for similar features in subsequent frames, and iteratively achieve target positioning step by step, regardless of background information | (Zhou et al., 2009) |
| | Discriminative | Consider the background information and target model, and detect the current frame position of the target through difference comparison | (F. Li et al., 2018) |
| **Model Fusion** | TBD | Take the current state of the relevant target as input, and use the tracking algorithm to predict the position | (Wojke et al., 2017) |
| | JDE | Combine the two parts of detection and recognition into a first-level network | (Li et al., 2022) |

# 3 Components of the Fish Tracking

Fish tracking methods can be categorized into two types based on the target initialization: manual marking of the target location, and automatic detection using recognition algorithms. The manual marking method is relatively fast but lacks the capability to detect and track new targets as they appear in the video sequence. The latter automatic detection method first identifies the target of interest within the data sequence, then directly matches and associates the detected targets across continuous frames. Fig. 5 provides a summary and classification of fish tracking methods. The majority of researchers continue to utilize the classic TBD method. This section primarily compiles algorithms from both the position initialization and tracking stages under TBD framework.

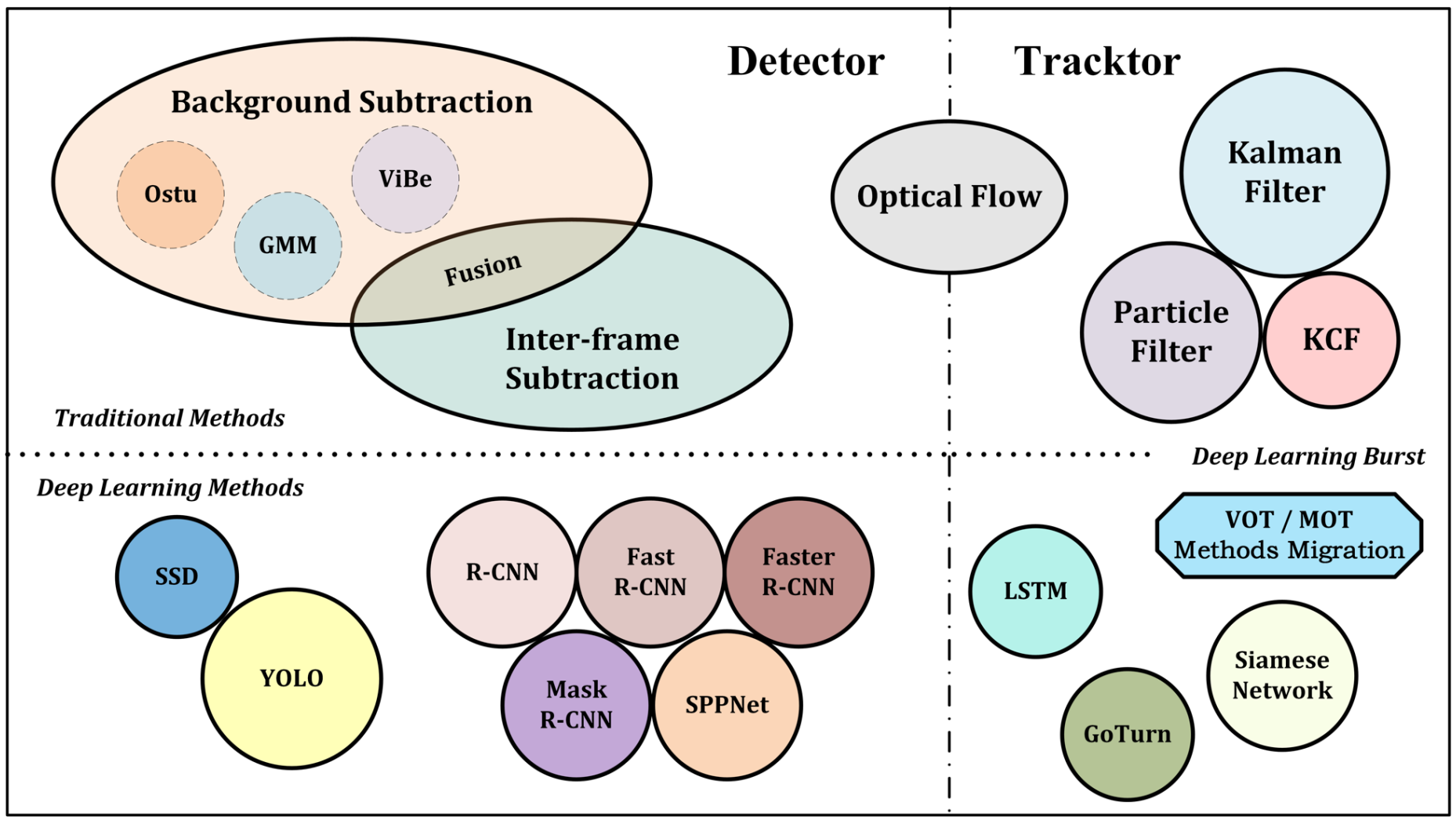

Fig. 5 Typical fish tracking methods and detector (initialization) components.

## 3.1 Fish Detectors

Fish tracking places greater emphasis on edge features and contextual information. The performance of the detector (or initialization method) significantly influences the quality of fish tracking results. Due to the typically poor illumination and clarity of underwater videos and images compared to general data, background subtraction is commonly used as a method for extracting fish targets. It can yield satisfactory results for data captured by static cameras. For traditional fish detection or initialization, many methods utilize background subtraction or its variants due to the simplicity and efficiency of these techniques. For instance, Zhao et al. (2019) introduced an approach using Otsu adaptive segmentation and interframe relationship. In contrast to basic thresholding, Otsu provides a quick, unsupervised method for fish positioning and feature extraction. However, its performance decreases when water quality is poor or fish overlap significantly. Shevchenko et al. (2018) compared three background subtraction methods to handle different fish movements. Their experiments showed Adaptive Gaussian Mixture Model (GMM) and Visual Background Extractor (ViBe) achieved satisfactory results. In general, background modeling is effective for slow motions and robust for complex backgrounds. However, challenges arise in noisy and cluttered conditions.

Traditional fish tracking methods are often considered less robust due to their limited region selection strategies and reliance on manual feature extraction. In contrast, deep learning methods are progressively replacing these conventional approaches, owing to their powerful learning capabilities, adaptability, and data-driven nature. A prevalent approach involves the adoption of detectors from the You Only Look Once (YOLO) series for extracting positional and feature information (Banerjee et al., 2021; Liu et al., 2021; Mohamed et al., 2020). Compared to alternative algorithms, YOLO has several advantages, including high-speed processing enabling real-time detection, considering the full image as context to reduce background misclassification, and robust generalization capability. YOLO can be optimized end-to-end to enhance detection performance. Liu et al. (2018) combined YOLOv3 with parallel correlation filters for online fish detection and tracking. YOLOv3 uses logistic regression to predict scores for each bounding box. Although multi-label classification can predict potential categories, it may not effectively handle occluded fish. Mohamed et al. (2020) presented a fish trajectory detection technology based on YOLOv3. They combined YOLO with optical flow to capture motion trajectories of multiple fish. This resulted in improved detection accuracy, especially in turbid water. Liu et al. (2021) proposed a real-time fish detection and tracking method using YOLOv4, which can generate population statistics for various species. This enhances performance in complex underwater environments. In summary, YOLO-based detectors have consistently demonstrated excellent target localization across diverse scenes and species, exhibiting robustness and promising development potential.

## 3.2 Fish Tracktors

### 3.2.1 Heuristic-based Tracktors

***Filter.*** Fish tracking models typically adopt kinematic methods with filter-based tracking, including Kalman filter, particle filter, and kernel correlation filter. The Kalman filter is a popular and effective method for system state estimation. It excels at making informed predictions in dynamic systems with uncertain information. A single Kalman filter is used to track the center of each fish, but it may be significantly affected under conditions of high turbulence (Zhao et al., 2019). The particle filter adopts nonlinear, non-Gaussian motion models for robustness when dealing with complex movements. It uses adaptive partitioning and nearest neighbor data association to analyze fish trajectories. However, there remains potential to further improve the robustness of this method. The kernel correlation filter is a discriminant tracking method that demonstrates fast and accurate performance. A primary limitation of filter-based approaches is the requirement for large sample sizes to effectively approximate posterior probability densities (X. Li et al., 2018).

***Optical Flow.*** Another approach uses optical flow, which refers to the instant velocity of a moving object derived from pixel changes on the image plane. It leverages pixel variations over time and inter-frame correlations to establish correspondence across frames for tracking. Optical flow is widely used for motion information in detection tasks. For example, Mohamed et al. (2020) applied optical flow to recognize fish movement, by passing the center point from object detection frames to enable tracking. They also combined Retinex color enhancement with YOLO to mitigate the impact of turbid water. Banerjee et al. (2021) proposed fish tracking using dense optical flow to estimate inter-frame pixel displacement. They observed polynomial motion transformations to define a method for estimating the displacement field from polynomial expansion coefficients. A detection algorithm was used to obtain frame masks representing fish positions, facilitating optical flow tracking. However, it should be noted that optical flow-based tracking relies heavily on detector performance and is much more computationally demanding and time-sensitive.

***SORT.*** The SORT algorithms (Bewley et al., 2016; Wojke et al., 2017) are widely used in MOT for their simplicity and strong performance. Compared to SORT, DeepSORT incorporates an appearance model for long-term tracking, partially addressing target identity loss. These trackers use Kalman filter for state prediction and the Hungarian algorithm to calculate instance similarity based on detection frame position and Intersection over Union (IoU). In essence, they integrate filter and optimal assignment into a MOT framework. For example, Kay et al. (2022) adopted SORT as the baseline for proposed fish tracking benchmarks. Experiments showed this SORT-based method performed comparably to human experts on their dataset, with under 5% error rate. These trackers achieve excellent performance at high frame rates, though overall accuracy drops with lower sub-model accuracy. Challenges remain in network sharing across stages and further speed improvements. In general, SORT and its variants have achieved considerable success due to strong overall performance for fish tracking.

### 3.2.2 Learning-based Tracktors

***LSTM.*** Long Short-Term Memory (LSTM) is a recurrent neural network that can effectively mitigate the problem of vanishing or exploding gradients during training on long sequences. Palconit et al. (2020) proposed a model that utilizes genetic algorithms and LSTM to predict fish trajectories. It tentatively uses genetic algorithms to predict the shortest path of fish, achieving promising results. Detection is based on background subtraction and blob analysis to extract information such as the bounding box, location, and area of the fish target. The genetic algorithm employs linear regression as the fitness function and utilizes tournament selection to find the optimal coordinates and select the best solution. In the LSTM network, it takes the detected centroid positions of the fish target in three consecutive frames as input for trajectory prediction. Gupta et al. (2021) utilized a fusion of LSTM with an attention mechanism to enhance the accuracy of bounding box prediction. They argued that not all bounding boxes on each fish trajectory contribute equally to the learned features. The motion trajectories of fish are mostly streamlined rather than highly chaotic. By introducing an attention mechanism in LSTM, they achieved the acquisition of context vectors and assigned more weight to bounding boxes with abrupt changes in their motion trajectories. While these methods have shown good performance in processing video data under turbid waters, they do not adequately incorporate the informative supervision from the annotations. In other words, better utilization of annotation information could help further improve model performance.

***GOTURN.*** GOTURN is an offline tracker based on deep learning methods, known for its high processing speed. It utilizes simple forward propagation to achieve real-time tracking speeds of up to 100 frames per second. This approach effectively prevents overfitting, as the model learns generalized motion features that can be applied to effectively track new objects without prior observations. In complex underwater environments, traditional tracking methods often suffer from high false negative rates, frequently resulting in failures in fish tracking tasks. GOTURN demonstrates potential to overcome these challenges through its efficient deep learning framework and generalized tracking capabilities. Arvind et al. (2019) conducted a comparative study on fish detection and tracking using high-resolution aerial image data from multiple regions. They employed Mask R-CNN for real-time fish detection to generate candidate regions, followed by GOTURN tracking to trace candidate trajectories. This approach runs the detection model on imagery captured by a drone, creates and classifies fish masks, and leverages the efficiency of GOTURN for real-time tracking and counting. It demonstrates improved computational efficiency and enables real-time analysis. However, in turbid water conditions, underwater visibility is reduced, which can result in failures to capture clear aerial imagery. This may lead to missed detections as unmanned aerial vehicles struggle to acquire high-quality data in turbid environments. Overall, while the approach shows promise, performance may be limited by water clarity when applied to turbid conditions.

***Siamese Network.*** Siamese network-based trackers have received considerable attention recently due to their ability to achieve a favorable balance between tracking accuracy and computational efficiency (B. Li et al., 2019). They transform the target tracking problem into a patch block matching task, as illustrated in the development timeline shown in Fig. 6. For instance, SiamMOT (Shuai et al., 2021) employs area-based trackers to model instance-level motion. However, practical applications of Siamese networks in fish tracking face challenges due to the difficulties in distinguishing characteristics among different fish instances. Shen et al. (2022) introduced ULAST for unsupervised learning tracking. They designed a differentiable region mask to select features and implicitly penalize tracking errors in intermediate frames. In fish tracking, Wang et al. (2022) achieved multiple fish tracking and abnormal behavior detection using SiamRPN++ and YOLOv5. SiamRPN++ formulates the target tracking problem as a cross-correlation task and utilizes ResNet-50 as a deep backbone network to extract color, shape, and location cues at lower layers as well as more semantic information at higher layers. Siamese networks can achieve performance comparable to filter-based methods and hold great promise for data-driven research. However, it is important to note that Siamese networks do not natively support MOT and require substantial GPU resources.

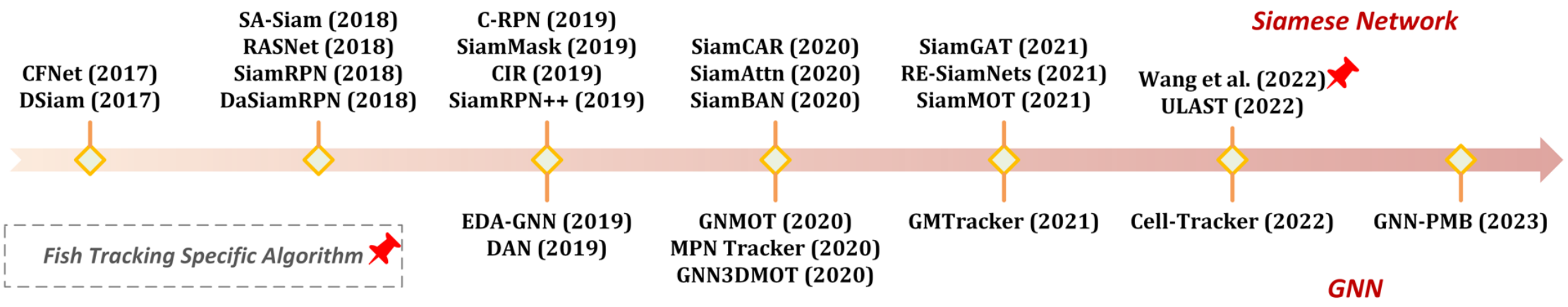

Fig. 6 Siamese networks and GNNs in the past 7 years (2017-2023).

***GNN.*** Graph Neural Network (GNN) are finding applications in tracking due to their strength in modeling relationships between objects. Li et al. (2020) designed a nearly online MOT method using GNN. They introduced a missed detection strategy to notify the detector of any defects, and incorporated an update mechanism for nodes, edges, and node variables. Jiang et al. (2019) developed an end-to-end framework combining affinity learning and optimization with GNN to address data association challenges in online tracking. Braso et al. (2020) employed a classical network flow formulation of MOT to establish a fully differentiable framework based on Message Passing Network (MPN). This approach conducts global inference on the entire detection set and predicts the final solution. Dai et al. (2021) designed a proposal-based learning framework for MOT, which divides the process into three steps: generating proposals using iterative graph clustering, scoring proposals and learning structural patterns using graph convolutional networks, and inferring trajectories on an affinity graph. This framework improves proposal quality, reduces computation costs, and enhances prediction accuracy. In summary, GNN excel at extracting associations between instances and can effectively enhance tracking accuracy in MOT. As shown in Fig. 6, the development momentum of GNN has been observed in other domains, but their application to fish tracking remains limited due to high computational costs.

***Transformer.*** Transformers have achieved success in various computer vision tasks, including image recognition, object detection, segmentation, and super-resolution, primarily owing to their self-attention mechanism (Khan et al., 2022). One notable application is in the field of MOT, where the Transformer architecture has been adopted. For example, TransTrack (Sun et al., 2021) adopts the JDE paradigm and introduces the Transformer to MOT for the first time. It utilizes the query-key mechanism to design a detection branch that queries target positions from keys. In the tracking branch, it queries the target positions in the current frame from keys to obtain tracking boxes, followed by IoU matching to complete tracking. This approach provides faster processing speed and incorporates a guided Transformer encoder module that considers global semantic relations often overlooked. MOTR (Zeng et al., 2022) has designed an end-to-end transformer-based MOT framework that enables overall network architecture training. Subsequent versions of the model, including MOTRv2 and MOTRv3, have further explored optimized detector incorporation and model training mechanisms. Transformer-based models have the capability to capture long-range relational dependencies among fish targets, similar to GNN architectures. However, they are not yet widely adopted for fish target tracking due to the high computational cost associated with network training.

# 4 Auxiliary Plugins

## 4.1 Underwater Image Enhancement

Underwater image degradation severely affects position initialization and stability of fish target tracking, which can potentially be addressed by exploring image enhancement and restoration techniques (Liu et al., 2021; Mohamed et al., 2020; Wageeh et al., 2021). Obtaining fish data in deep underwater environments poses difficulties due to factors such as light attenuation and underwater noise. The labor-intensive requirements for manual labeling also make it challenging to analyze original videos and extract meaningful data. Leveraging underwater image enhancement technologies can substantially improve the quality of degraded visual data, thereby facilitating the development of new fish tracking methods.

Underwater images often suffer from quality degradation due to light absorption and scattering in water. General image enhancement methods may not perform well in processing underwater data. There are two broad categories of underwater image enhancement methods: *Traditional methods* often use physical models, such as atmospheric scattering models, simplified underwater imaging models, and modified underwater imaging models. They take into account various factors affecting image quality and aim to enhance individual pixel values in underwater images. The other category is *Deep learning-based methods*, such as Deep Convolutional Neural Network (Deep-CNN) and Generative Adversarial Networks (GAN). CNN-based methods focus on preserving the original images, while GAN-based methods aim to improve image quality. Both traditional and deep learning-based methods play crucial roles in addressing underwater imaging challenges, enhancing image quality, and various applications, including fish target tracking in underwater environments.

**Table 2.** Deep underwater image enhancement networks.

| Network | Example |
|---|---|
| Encoder-Decoder Networks | P2P (Sun et al., 2019), UGAN (Fabbri et al., 2018) |
| Modular Design Networks | UWCNN (Anwar et al., 2018), DenseGAN (Guo et al., 2020) |
| Multi-Branch Designs | DUIENet (C. Li et al., 2020), FGAN (H. Li et al., 2019) |
| Depth-Guided Networks | URCNN (Hou et al., 2018), WaterGAN (Li et al., 2017) |
| Dual Generator GANs | UWGAN (C. Li et al., 2018), MCycleGAN (Lu et al., 2019) |

Table 2 provides examples of deep networks used for underwater data enhancement, each with its specific goal and approach. These techniques are essential for enhancing the quality of underwater images, thus improving the accuracy of fish target tracking in challenging underwater environments.

## 4.2 Fish Re-identification

Appearance models are valuable tools for achieving individual identity confirmation of targets and extracting appearance features in fish tracking, helping address issues arising from fish occlusion. Additionally, Re-identification (Re-ID) methods, which are popular in computer vision, find applications in fish tracking by leveraging appearance features. These Re-ID models are embedded as sub-modules within fish tracking networks to enhance tracking performance, as illustrated in Fig. 7. For example, Li et al. (2018) developed a fish tracking model using Re-ID, focusing on appearance features. This method models multiple appearances of normal and abnormal fish bodies. DFTNet (Gupta et al., 2021) employed a Siamese Network to extract appearance features for differentiating fish targets based on the number of fish being tracked. In some novel JDE paradigms, appearance feature extraction is integrated into the entire network architecture. These end-to-end methods have also been applied to fish targets. For example, CMFTNet (Li et al., 2022) proposed a multiple fish tracking network based on the JDE paradigm, embedding the Re-ID module as a network branch. This allows online tracking while leveraging the learned embeddings to improve segmentation mask quality. Generally speaking, these Re-ID embedding-based models are expected to further evolve and optimize tracking performance.

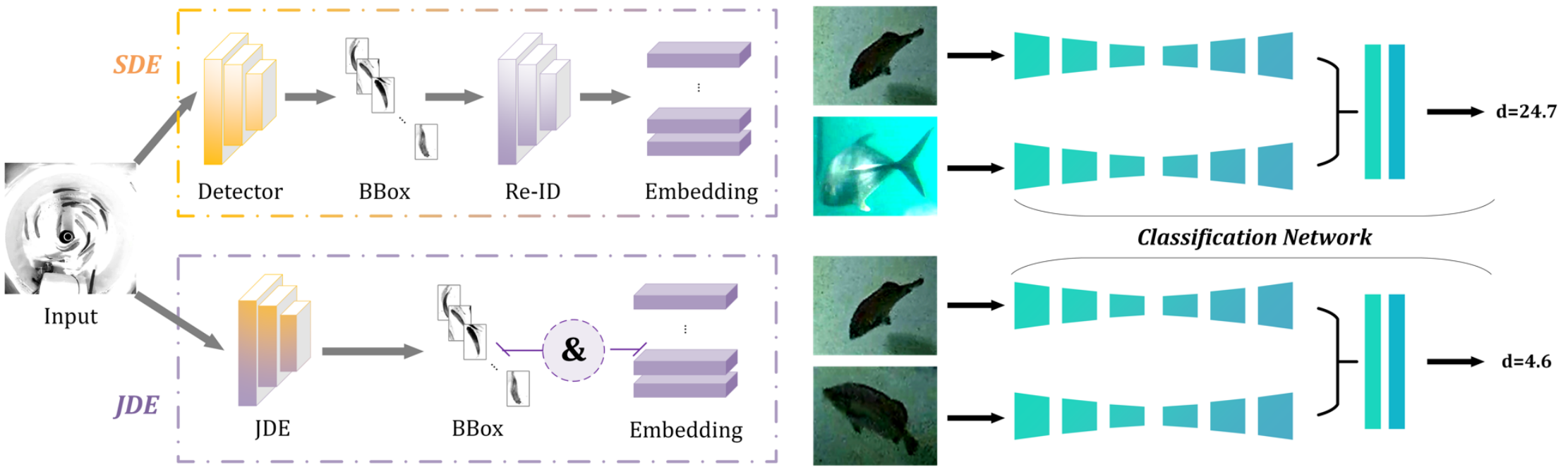

(a) Re-ID Embedding in SDE and JDE Paradigms (b) Appearance Feature Extraction in Re-ID

**Fig. 7** Examples of various tracking paradigms and appearance feature extraction in Re-ID. Different colors in the figure represent different semantic information, e.g., (a) yellow represents the detection box and purple represents ID information in the figure.

# 5 Datasets and Metrics

## 5.1 Open-Source Fish Datasets

**Table 3.** Comparison of several typical open-source fish datasets.

| Dataset | Fish4Knowledge | LifeCLEF2014 | LifeCLEF2015 | Labeled Fishes in the Wild |
|---|---|---|---|---|
| **Type** | Video / Image | Video | Video / Image | Video / Image |
| **Volume** | 200TB, 6,514 video clips / 27,370 images, 23 types | About 1,000 video clips | 20 videos / 20,000 images | An ROV video, 4096 images |
| **Description** | The data categories collected are relatively unbalanced. | From the Fish4Knowledge video dataset, including algae attachment data and turbid water data, difficult to identify. | Both video and image data are clearly labeled, and there is still a problem of imbalance of data samples. | Contains positive and negative datasets. |
| **Website** | *https://groups.inf.ed.ac.uk/f4k/* | *https://www.imageclef.org/2014/lifeclef/fish* | *https://www.imageclef.org/lifeclef/2015/fish* | *https://swfscdata.nmfs.noaa.gov/labeled-fishes-in-the-wild/* |

This section provides an overview of the relevant underwater datasets and processing methods for developing and evaluating fish tracking algorithms. The annotation information such as location and bounding box in the dataset can be utilized to accomplish detector training and visualization of the fish tracking model. Notably, this type of detection requires valid ID labeling in order to acquire tracking metrics. Several representative datasets for fish tracking research have been compiled, as shown in Table 3.

***Fish4Knowledge.*** The Fish4Knowledge (F4K) dataset is a large-scale open-source dataset funded by organizations including the European Union's Seventh Framework Program and the Taiwan Power Company of China. As shown in Fig. 8a, the F4K dataset contains video frames captured from underwater scenes. The project's primarily focus is on methods for efficient video capture, storage, analysis and querying, with the data collected enabling fish detection and tracking research. In the F4K project, 10 cameras recorded over 12 hours of footage daily, amassing approximately 100TB of data annually. F4K is one of the largest public datasets for fish tracking, comprising a total of 23 fish species categories. The scale and diversity of F4K make it a highly valuable resource to spur advances in fish tracking techniques. However, it is important to note that the dataset exhibits considerable class imbalance among the different species.

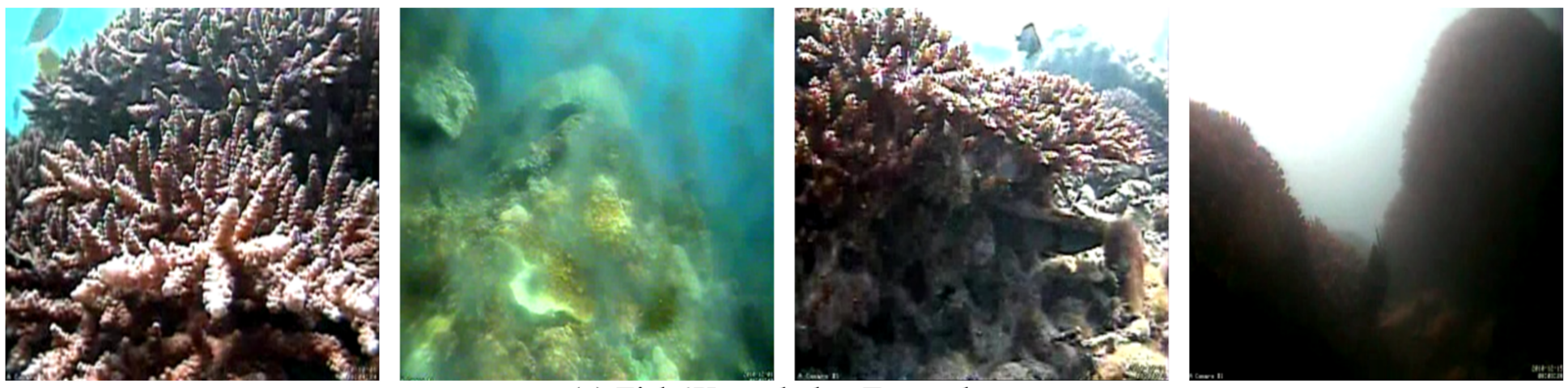

(a) Fish4Knowledge Examples

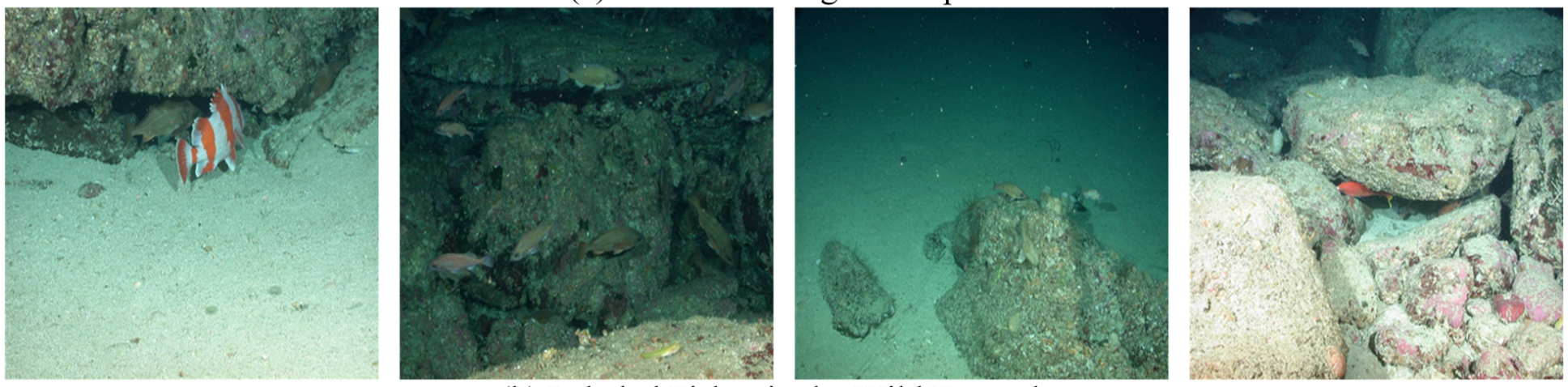

(b) Labeled Fishes in the Wild Examples

**Fig. 8** Open-source datasets samples.

***LifeCLEF.*** The LifeCLEF initiative has developed comprehensive biological databases spanning species taxonomy, geography, and evolution. Over the years, LifeCLEF has released several fish and marine life datasets, including FishCLEF2014, FishCLEF2015, SeaCLEF2016, and SeaCLEF2017. The FishCLEF2014 underwater video dataset is derived from the Fish4Knowledge database, comprising around 700,000 10-minute clips captured over 5 years in Taiwan's coral reefs. The videos cover a wide temporal range from sunrise to sunset. Challenges arise from the turbid water, camera lens artifacts, and algae growth on the lenses that obstruct fish identification. The dataset has a 40GB training set and 8GB test set with four sub-datasets for different tasks. One focuses on image-based species classification, while the others contain videos. Similar to FishCLEF2014, FishCLEF2015 utilizes Fish4Knowledge data. The training set has 20 annotated videos, a species list, and over 20,000 sample images. Testing data has 73 videos and a species list, facing class imbalance challenges. Building on the FishCLEF fish identification tasks, SeaCLEF2016 and SeaCLEF2017 expand the scope to broader marine life. By moving beyond just the Fish4Knowledge source, these datasets enable improved model training compared to the previous FishCLEF releases.

***Labeled Fishes in the Wild.*** The Labeled Fishes in the Wild dataset was developed by NOAA Fisheries (National Marine Fisheries Service) to promote research on unconstrained automatic image analysis algorithms for fish identification. The dataset primarily comprises images of fish, invertebrates, seabed, and annotation files specifying the locations of fish targets. It includes several key components: a positive training/validation set with fish images, a negative training/validation set without fish, and a testing image set. For both training and testing, the locations and sizes of fish targets are annotated in each image. The images were captured by a forward-facing digital camera mounted on a Remote Operated Vehicle (ROV) deployed by the Southwest Fisheries Science Center. The positive set contains 929 images and 1,005 annotations, while the negative set has 3,167 images. The testing set utilizes ROV HD video sequences, with 2,061 annotated fish objects in total. Owing to its high-quality images, clear organization, and comprehensive annotations, this dataset represents a valuable resource to spur advances in fish tracking research and development.

## 5.2 Metrics

**Table 4.** MOT evaluation metrics.

| Metric | | Description | Better |
|---|---|---|---|
| **Comprehensive Metric** | **MOTA** | Multiple Object Tracking Accuracy, involving false positives, missed targets and identity switches. | Higher ↑ |
| | **IDF1** | The ratio of correctly identified detections over the average number of ground-truth and computed detections. | Higher ↑ |
| | **HOTA** | Higher Order Tracking Accuracy. Geometric mean of detection accuracy and association accuracy. | Higher ↑ |
| **Detailed Metric** | **MT** | Mostly Tracked targets. The trajectory prediction covers those ground-truth accounts for more than 80% of the total. | Higher ↑ |
| | **ML** | Mostly Lost targets. The trajectory prediction covers those ground-truth accounts for less than 20% of the total. | Lower ↓ |
| | **Rcll** | Ratio of correct detections to total number of ground-truth. | Higher ↑ |
| | **ID Sw.** | Number of Identity Switches. | Lower ↓ |

In the context of fish tracking, the emphasis primarily lies in achieving simultaneous tracking of multiple fish, making MOT metrics the prevalent choice for evaluation. This subsection provides a concise summary of the evaluation indicators commonly adopted in MOT. The MOT Challenge has been instrumental in this regard, offering multi-target pedestrian datasets, with the MOT16, MOT17, and MOT20 datasets serving as the most widely adopted benchmarks. Despite being initially designed for pedestrian tracking, these datasets have found application in evaluating fish tracking algorithms due to the shared principles of MOT. They provide real-world scenarios and meticulously annotated ground truth data, enabling comprehensive assessments of fish tracking algorithms. Although there are distinctions between pedestrian and fish tracking, these MOT evaluation metrics can effectively measure the performance of fish tracking algorithms in multi-target scenarios. The typical evaluation metrics used for MOT algorithms include the CLEAR (Bernardin and Stiefelhagen, 2008) and HOTA (Luiten et al., 2021) metrics. These metrics are detailed in Table 4 for reference. It is worth noting that MOTA, IDF1, and HOTA are comprehensive evaluation metrics commonly employed in most MOT tasks.

***MOTA.*** Multiple Object Tracking Accuracy (MOTA) is a metric used to measure the tracking accuracy of single-camera multi-target tracking. It is calculated using the following formula:

$$MOTA = 1 - \sum(FN + FP + ID_{Sw})/\sum GT \quad (1)$$

$FN$ is false negative, $FP$ is false positive, $ID_{Sw}$ is identification switch, and $GT$ is the number of ground truth objects. MOTA is a valuable metric for evaluating tracking algorithms as it considers missed detections, false alarms, and identity switches, providing a comprehensive assessment of tracking accuracy. It focuses on the accuracy of detection and tracking. A higher MOTA value indicates better tracking performance. However, MOTA cannot reflect the continuity of long-term tracking very well.

***IDF1.*** Identification-Score (IDF1) is a comprehensive metric that considers both Identification Precision (IDP) and Identification Recall (IDR). The formula for IDF1 is:

$$IDF_1 = TP/(TP + 0.5FP + 0.5FN) \quad (2)$$

$TP$ is true positive. IDF1 effectively evaluates the ability of a tracking algorithm to correctly identify trajectories that correspond to the same target. IDF1 focuses on the ability to continuously track the same object correctly. A higher IDF1 means better tracking continuity. However, IDF1 cannot directly reflect the accuracy of detection and matching.

***HOTA.*** Higher Order Tracking Accuracy (HOTA) (Luiten et al., 2021) is a metric designed to better align evaluation scores with human visual assessments. The formulas for calculating HOTA are:

$$HOTA_\alpha = \sqrt{\Sigma_{c\epsilon\{TP\}} A(c)/(|TP| + |FN| + |FP|)} \quad (3)$$

$$A(c) = |TPA(c)|/(|TPA(c)| + |FNA(c)| + |FPA(c)|) \quad (4)$$

$\alpha$ is the IoU threshold and $c$ is the number of positive sample trajectories. The intersection between two trajectories is termed a True Positive Association (TPA), while the trajectories outside the intersection in the predicted trajectory are termed False Positive Associations (FPA). Detections outside the intersection in the ground-truth trajectory are termed False Negative Associations (FNA). HOTA effectively addresses issues related to overemphasizing detection or association. It has become a widely accepted and interpretable evaluation metric for MOT.

# 6 Challenges and Applications

## 6.1 Challenges

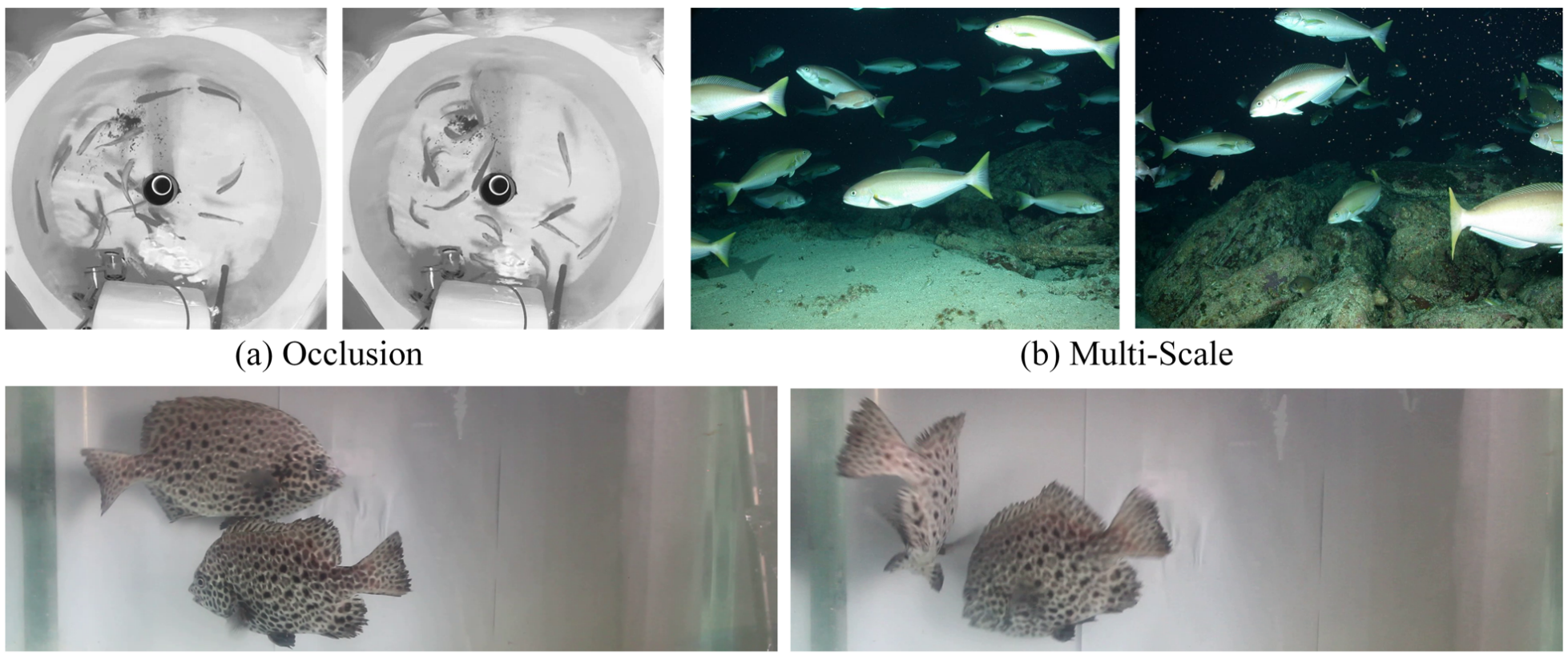

(a) Occlusion (b) Multi-Scale

(c) Deformation

Fig. 9 Challenges of fish tracking.

Significant challenges in fish target tracking include occlusion, multi-scale, and deformation phenomena. These challenges arise from the camera's positioning and the natural movement patterns of fish, as illustrated by the example images in Fig. 9.

### 6.1.1 Occlusion

In practical applications, one solution to address occlusion problems in fish target tracking is to use a system as shown in Fig. 10. This typically involves either a 3D tracking solution or an additional step for multi-view alignment. Current methods often collect data using such an acquisition setup, where the viewpoints with less occlusion are selected as the core dataset for model design. Multiple cameras are strategically positioned around the fish tank to capture synchronized videos of fish behavior. The data is then synchronized to a computer for further processing.

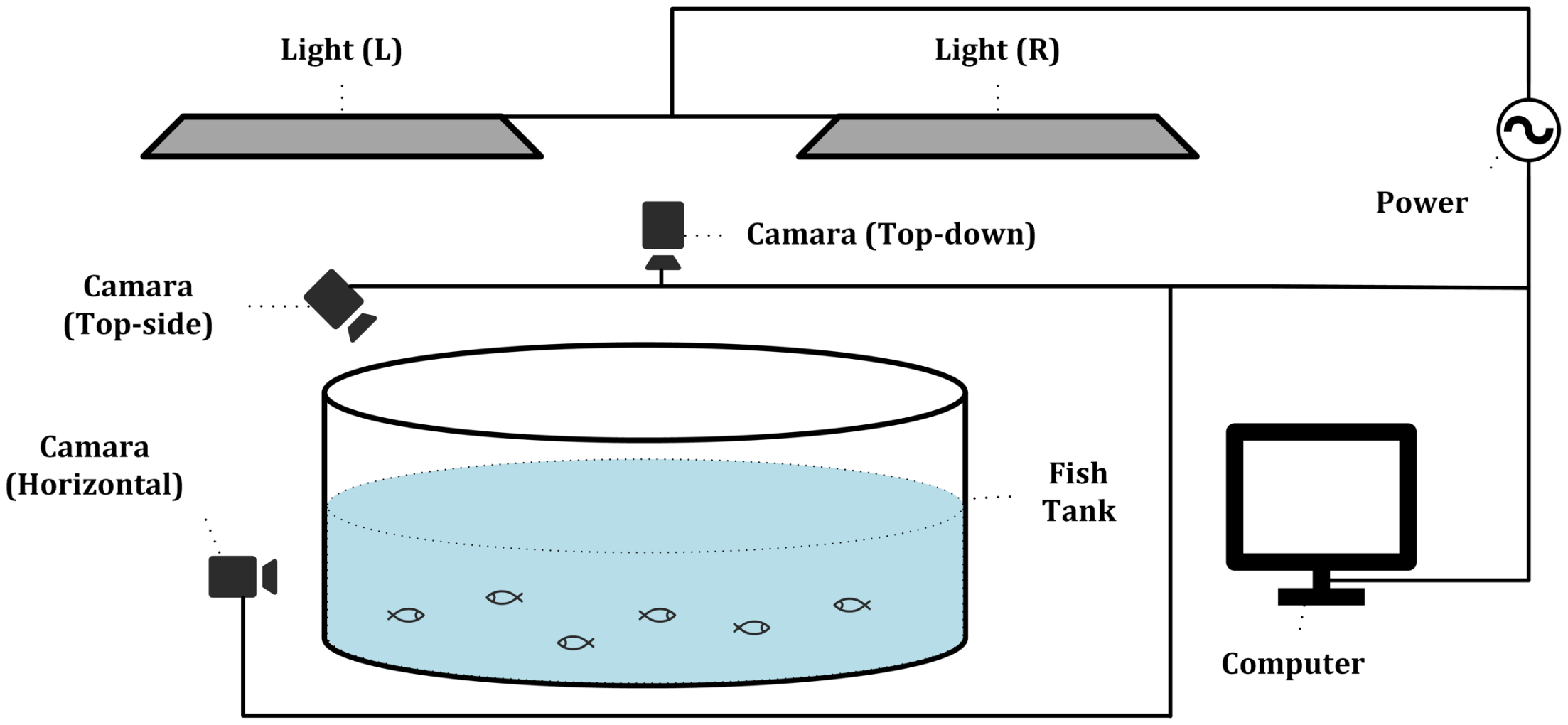

**Fig. 10** Application of multi-view fish culture pond image acquisition.

To address these challenges, common methods involve employing master-slave camera setups or using cameras with mirrors, among other techniques. For instance, Liu et al. (2019) introduced a 3D tracking method based on fish skeletons. In this approach, top and side views are utilized to simplify the fish target into a representation of feature points. These points are then associated with the movement trajectory observed from the top view. The continuity constraint matches the feature points from both views, ultimately providing a 3D trajectory. However, this method has limitations, particularly its reliance on an asymmetric strip structure. It can also be complicated in terms of calculations and equipment setup when using master-slave cameras or mirrors. As a result, many researchers prefer to predict and estimate fish trajectories from single-camera video data.

### 6.1.2 Multi-Scale and Deformation

Another challenging issue in fish target tracking is the high-frequency transformation of scale and geometry of the fish targets. Addressing this requires careful consideration in network architecture design. One approach is to leverage deep learning-based networks that can effectively handle scale-related challenges. For example, the Feature Pyramid Network (FPN) combines deep feature maps with strong semantic information and shallow feature maps with high resolution. Additionally, the Deformable Convolution Network (DCN) can help address the problem of target deformation arising from scale transformations and the non-rigid characteristics of fish, as illustrated in Fig. 11. Anchor-free detectors provide solutions to overcome the challenges posed by deformation and scale transformations to some degree. For instance, Zhou et al. (2020) have adopted an anchor-free detection method, utilizing heatmaps to localize the object's center. This exhibits superior robustness across various target scales. Their model strikes a balance between speed and accuracy, making it suitable for integration into 3D applications. The anchor-free tracking approach mitigates frequent scale variations, reducing matching errors related to overlap. It can also effectively identify high-density distributions, aiding the detection of behaviors like feeding and alarm responses. Nian et al. (2017) proposed a portable smart device with an online fish detection and tracking strategy, employing a combination of deformable fish body representations and compressive sensing techniques. The fish detection system uses a mixture of multi-scale deformable models to represent highly variable underwater objects, capturing fine-resolution features of fish through a fish body part filter.

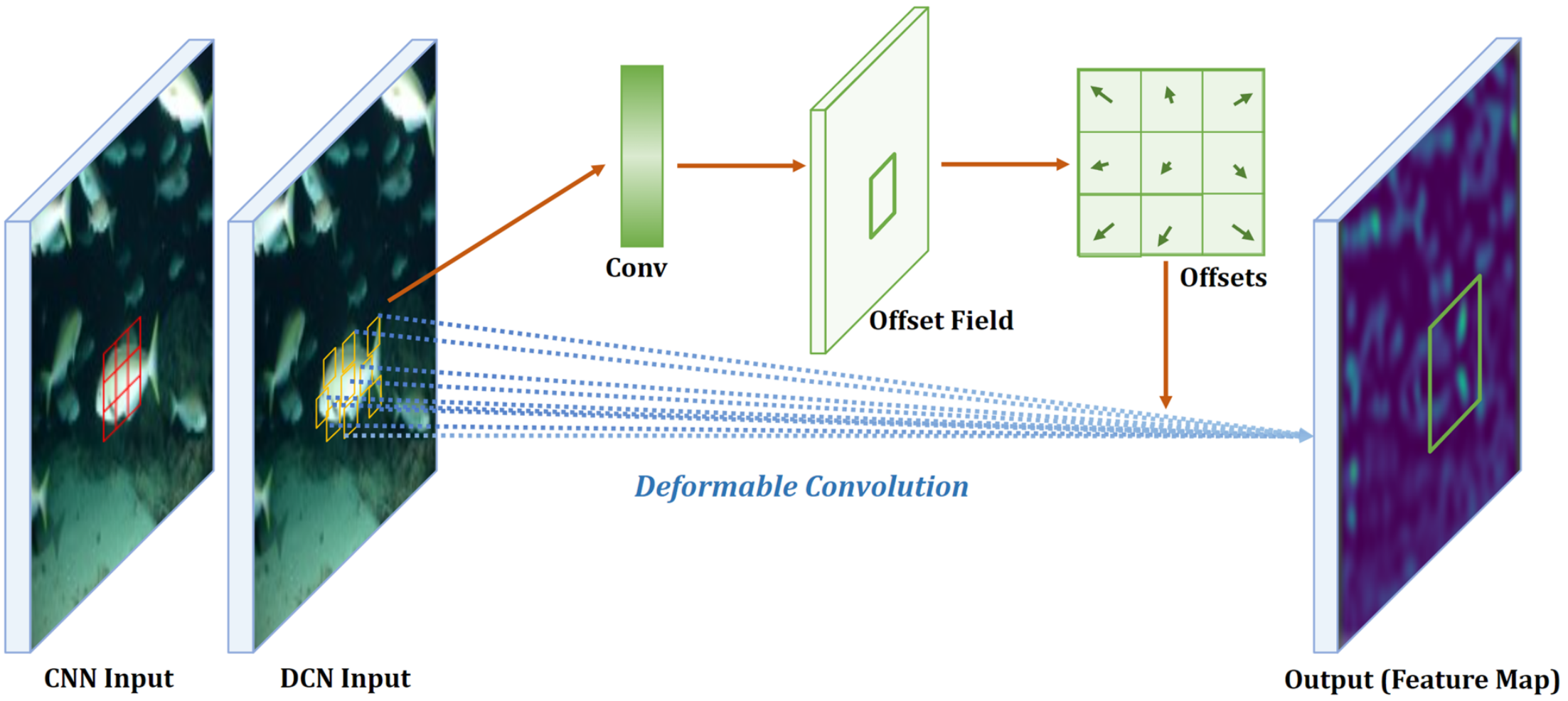

Fig. 11 Deformable convolution network (DCN).

## 6.2 Applications

The tracking methods ultimately generate trajectory data for individual fish, recording their positions in each frame. By analyzing fish movement trajectories, extracting embedded information, and leveraging the synergy of hardware and multidisciplinary knowledge, it becomes feasible to undertake various multi-application tasks. These include fish behavior analysis, dynamic fish counting, and marine ecological monitoring.

When integrated with biological domain knowledge, fish tracking can be utilized to identify abnormal fish behavior trajectories. The key challenge is developing an abnormality evaluation model that can assess whether fish are exhibiting unusual behaviors related to factors such as hunger, parasites, or hypoxia. For example, Shreesha et al. (2020) investigated three behavioral patterns in Sillago sihama and developed a decision support system based on motion data from the tracking model. However, this model is complex and requires a larger training dataset to improve accuracy. Similarly, Wang et al. (2022) introduced a fish tracking method using YOLO and Siamese Network, which enabled identification and continuous localization of anomalous fish species. In their method, a YOLO detection algorithm is initially utilized for specific target differentiation, reducing the computational load of the tracking model.

Dynamic fish target counting is enabled through multiple fish tracking methods, effectively tackling issues related to repetitive counting in static methods. Static fish counting typically relies on capturing keyframes, but when fish exhibit varying movement speeds, the fixed keyframe capture frequency can lead to duplicate detections. Video tracking-based counting schemes assign different ID numbers to fish targets to eliminate duplicates. For example, França Albuquerque et al. (2019) proposed a live fingerling counting method and dataset based on fish tracking algorithms. Fish are counted by combining information from round block detection, Gaussian blending, and Kalman filter. This automated fish counting method can reduce costs, enhance production, and improve labor efficiency, achieving an average accuracy of 97.5%.

Fish possess highly sensitive sensory mechanisms, particularly olfaction, gustation, and thermoception. They also demonstrate sensitivity to variations in water quality parameters including salinity, pH, and dissolved oxygen concentrations. These factors can profoundly influence fish behavior, and their behavioral parameters are relatively facile to quantify. Abnormal fish behavior is frequently correlated with fluctuations in water quality. Analyzing fish behavior enables the derivation of characteristic variables responsive to changes in water quality, offering a valuable solution for water quality monitoring. Fish tracking techniques not only facilitate the tracking of fish but also furnish data support to monitor fish abundances, species, and behavioral traits within ocean observation networks. Nian et al. (2017) proposed a technique combining variable fish body representations with compressed sensing approaches. This methodology employs multi-scale variable fish body descriptions to represent mixed fish populations and entails the development of a portable ocean observation network device with online fish detection and tracking capabilities. The system effectively accomplishes real-time fish detection and multi-fish tracking. Experimental outcomes utilizing underwater video sequences demonstrate the efficiency and accuracy of the proposed fish detection and tracking strategy.

# 7 Discussion

Underwater environments present unique challenges for computer vision, including poor visibility, inter-object occlusion, non-rigid deformations, scale variations, and appearance changes due to lighting conditions and viewpoints. As a result, underwater fish tracking remains an open research area with significant potential for biological studies, aquaculture, and environmental monitoring. In this section, we have reviewed the primary challenges in underwater fish tracking from the perspectives of *data*, *paradigms*, *components* and *applications*.

***Data.*** The key challenges of fish tracking datasets relate to data acquisition and frame quality. Firstly, collecting fish datasets is exceptionally difficult due to the small size and high density of fish, making image labeling an arduous task. Existing open-source fish datasets often lack sufficient resolution and frame counts to meet the requirements of deep learning methods. As a result, there remains a shortage of high-quality open-source fish tracking datasets, leading many researchers to create proprietary datasets for experiments and evaluations. Secondly, fish target datasets frequently suffer from low brightness, low contrast, increased noise, and substantial color distortion. These factors impose significant obstacles for detection algorithms and further complicate developing accurate underwater fish tracking. Constructing a comprehensive, high-quality video dataset is critical for improving fish tracking algorithms. This requires integrating various techniques during data labeling, including classification, underwater image enhancement, and detection methods. Classification helps manage the dataset by dividing videos into scenarios and species, enabling balanced data distribution and training patterns. Underwater image enhancement corrects issues like low lighting and color shifts, using techniques such as diffusion-based enhancement to increase fish contour visibility and trajectory accuracy. Combining automated detection with manual correction ensures precise fish target annotation. This meticulous approach establishes a robust fish video dataset to support developing and evaluating fish tracking algorithms.

***Paradigms.*** The traditional two-step detached TBD tracking paradigm remains the dominant approach in fish tracking, despite the growing popularity of JDE models in MOT. Compared to the more recent JDE paradigm, the conventional TBD tracking framework continues to have advantages for fish tracking applications, enabling more comprehensive assessment of tracking accuracy and speed. The applicability of the TBD approach across diverse aquatic environments and species also contributes to its persistent prominence in this field.

***Components***. In the detection stage, traditional methods are often sufficient to meet localization needs for fish tracking, as fish have distinct shape and color features that contrast sharply with the water background. Common techniques used in this stage include basic background subtraction, inter-frame subtraction, or a combination of both. These techniques remain popular due to their computational simplicity. Recently, one-shot detectors based on deep learning have gained traction for real-time fish tracking applications, as they balance accuracy and speed in line with the demands of fishery production. These one-shot detection algorithms are being increasingly incorporated into the detection phase to enhance detection results for downstream trackers. Their development caters to the real-time performance requirements of fish tracking systems. When aiming for high-accuracy detection, they can be used as a pre-processing step alongside underwater image enhancement techniques or as integral components within network architectures. Similarly, in the context of the TBD paradigm, enhanced schemes, such as small target detection in general-purpose domains, can be adapted for fish target tracking.

In the tracking phase, kinematics-based filters remain prevalent in fish tracking. The well-known SORT algorithm remains integrated into most tracking models. While deep learning-based approaches have made progress in the detection phase, applying them to tracking remains an open challenge. Several researchers have explored data-driven models for inferring behavioral trajectories, but addressing the real-time nature of these approaches is a significant hurdle. Given traditional heuristic tracking methods have demonstrated robust performance in applications, they remain a focus of fish tracking research and enhancement. Tracking techniques developed for pedestrian and vehicle tracking, such as multi-level tracking, camera motion compensation, and multi-cue fusion, have shown excellent results and can also be leveraged for fish tracking. As for transformer-based tracking methods, they require large-scale fish datasets for optimal performance, while Siamese network-based tracking approaches struggle when dealing with large numbers of fish targets. From an application perspective, both tracking methods currently face challenges related to real-time processing and are more suitable for scientific research purposes.

***Applications.*** As previously discussed, different tracking schemes are employed for aquaculture and biological research applications, reflecting distinct tracking challenges. In aquaculture, multiple fish tracking methods are the predominant solution, primarily due to the prevalence of mutual occlusion, which significantly contributes to the loss of targets. The high-density distribution of fish often results in substantial mutual occlusion, making it considerably more challenging to extract individual fish features compared to re-identifying pedestrians in other contexts. Consequently, retrieving the original target becomes more intricate, leading to frequent recognition switches. To address short-term mutual occlusion issues, developing an optimized kinematics model for individual fish to facilitate recovering lost targets through algorithmic enhancements is essential. For long-term occlusion cases, obtaining detailed appearance differences among fish targets via close-up photography and high-definition data acquisition proves an effective approach. In biological studies, especially for fixed-point occlusion scenarios, employing master-slave cameras or cameras equipped with mirrors is recommended. These setups enable correlating multi-camera information and computing 3D coordinates, effectively mitigating target occlusion. By increasing computational complexity and capturing motion information in 3D space, this approach facilitates extracting spatial trajectories, thereby enhancing tracking accuracy during occlusion. Another practical challenge in fish tracking applications is the frequent acceleration of fish movement, often driven by foraging or sudden disturbances while feeding. This rapid acceleration can result in losing tracking targets. To effectively address target loss due to acceleration, recording video data at higher frame rates is a viable solution. However, this approach comes with the drawback of significantly increasing the computational load on hardware devices. Moreover, during model training, a higher number of frames can lead to reduced training efficiency. Typically, tracking models are trained using interval frames, and increasing the frames in such specialized scenarios can yield improved tracking results.

# 8 Conclusion

In this paper, we present a comprehensive survey of fish tracking technologies, encompassing various aspects including open-source fish datasets, preprocessing techniques, tracking methods, challenges, and applications. Our survey highlights that traditional subtraction and filter algorithms, grounded in a tracking-by-detection paradigm, have predominantly been employed in fish tracking. Furthermore, we observe the emergence of deep learning-based methods, which are progressively being applied to the detection stage, yielding notable improvements in tracking accuracy. Our compilation and analysis cover a broad spectrum of topics, including extant fish tracking paradigms, essential components within tracking systems, available fish tracking datasets, and associated evaluation metrics. Finally, we examine the prevailing bottlenecks and limitations in this field and offer insights into potential directions for future research.

# CRediT Authorship Contribution Statement

**Weiran Li:** Investigation, Resources, Writing - Original Draft, Writing - Review & Editing. **Zhenbo Li:** Formal analysis, Resources, Writing - Review & Editing, Supervision. **Fei Li:** Writing - Original Draft. **Meng Yuan:** Writing - Review & Editing. **Chaojun Cen:** Resources. **Yanyu Qi:** Writing - Review & Editing. **Qiannan Guo:** Investigation. **You Li:** Writing - Review & Editing.

# Declaration of Competing Interest

The authors declare that they have no known competing financial interests or personal relationships that could have appeared to influence the work reported in this paper.

# Acknowledgments

This study was supported by National Key R&D Program of China (2021ZD0113805); Key-Area Research and Development Program of Guangdong Province - Ecological engineering breeding technology and model in seawater ponds (2020B0202010009); National Key R&D Program of China - Integrated demonstration of seawater and brackish water fish ecological intensive intelligent breeding model and processing and circulation (2020YFD0900204).